\documentclass{article}
\usepackage{ijcai24}

\usepackage{times}
\usepackage{soul}
\usepackage{url}
\usepackage[hidelinks]{hyperref}
\usepackage[utf8]{inputenc}
\usepackage[small]{caption}
\usepackage{graphicx}
\usepackage{amsmath}
\usepackage{amssymb}
\usepackage{amsthm}
\usepackage{booktabs}
\usepackage{algorithm}
\usepackage{algorithmic}
\usepackage{booktabs} 
\usepackage{multirow}
\usepackage{arydshln}
\usepackage{comment}
\usepackage{siunitx} 
\usepackage[switch]{lineno}
\usepackage[marginal]{footmisc}

\title{MARS: Multimodal Active Robotic Sensing for Articulated Characterization}

\author{
Hongliang Zeng
\and
Ping Zhang\thanks{Corresponding author}
\and
Chengjiong Wu\and
Jiahua Wang\and
Tingyu Ye\And
Fang Li
\affiliations
South China University of Technology, Guangzhou, China
\emails
scutzenghongl@gmail.com,
pzhang@scut.edu.cn,
}

\begin{document}

\maketitle
\begin{abstract}
Precise perception of articulated objects is vital for empowering service robots. Recent studies mainly focus on point cloud, a single-modal approach, often neglecting vital texture and lighting details and assuming ideal conditions like optimal viewpoints, unrepresentative of real-world scenarios. To address these limitations, we introduce MARS, a novel framework for articulated object characterization. It features a multi-modal fusion module utilizing multi-scale RGB features to enhance point cloud features, coupled with reinforcement learning-based active sensing for autonomous optimization of observation viewpoints. In experiments conducted with various articulated object instances from the PartNet-Mobility dataset, our method outperformed current state-of-the-art methods in joint parameter estimation accuracy. Additionally, through active sensing, MARS further reduces errors, demonstrating enhanced efficiency in handling suboptimal viewpoints. Furthermore, our method effectively generalizes to real-world articulated objects, enhancing robot interactions. Code is available at https://github.com/robhlzeng/MARS.
\end{abstract}
\footnotetext{Copyright @2024,
	International Joint Conference on Artificial Intelligence (www.ijcai24.org).
	All rights reserved.}

\section{Introduction}
In an era increasingly marked by the integration of robotic assistance into everyday scenarios, conducting research on the precise perception of articulated objects, such as kitchen utensils and personal devices, is of paramount importance. These objects often have intricate joints and multiple moving parts, presenting unique and complex challenges in robotic perception and manipulation.

\begin{figure}[t]
	\centering
	\includegraphics[width=3.3in,keepaspectratio]{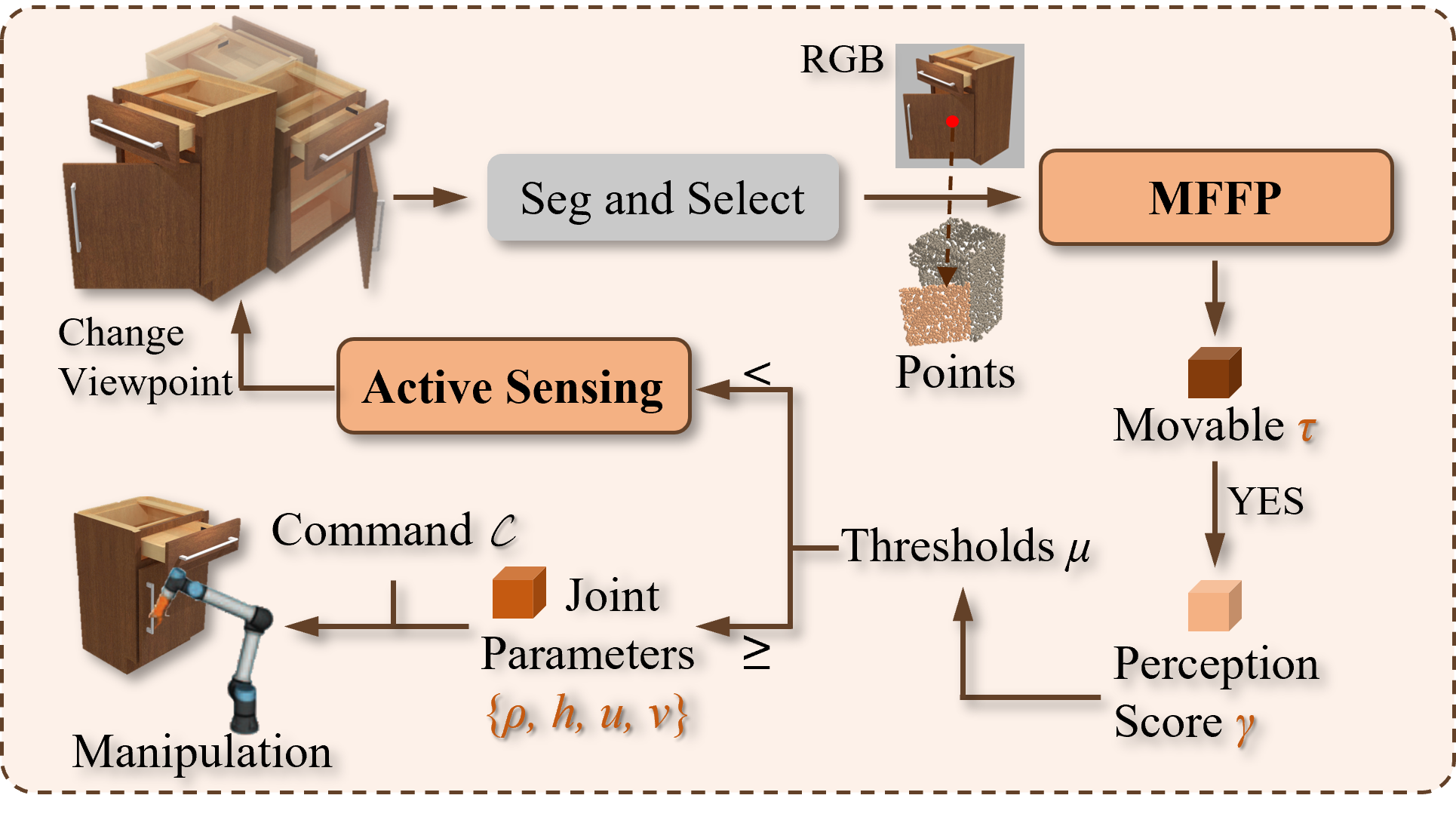}
	\caption{MARS uses active sensing to find optimal viewpoints for observing articulated objects, predicting precise joint parameters from RGB and point cloud for command-based robot planning. }
	\label{Graphical summary}
\end{figure}

By precisely perceiving the joint parameters of articulated parts, robots enhance their effectiveness in manipulation planning for these objects, considering parameters such as joint position, orientation, and part state. We reviewed previous research~\cite{yi2018deep,yan2020rpm,jain2021screwnet,jiang2022ditto,chu2023command} and identified limitations in existing methods. Firstly, Many studies perceive joint characteristics using a single point cloud modality, neglecting valuable information of color and texture data. Secondly, different joints, such as revolute and prismatic, necessitate separate perception networks, which can limit practical applications. Lastly, current work often assumes the availability of an ideal observation viewpoint, disregarding scenarios where the target part is obstructed or invisible. However, robots frequently face suboptimal viewing angles, hindering complete object observation.

To address these challenges, MARS implements a multimodal fusion strategy with a novel multi-layer dueling module, efficiently extracting and combining feature maps from diverse convolutional layers into a more effective, information-dense image feature representation. Subsequently, MARS integrates multi-scale RGB features and point cloud features via a transformer encoder-based fusion module. Moreover, MARS standardizes the description of joint parameters, allowing for the perception of various joint types through a single network. Addressing the issues posed by suboptimal viewing angles, MARS features a Reinforcement Learning (RL) driven active sensing strategy. This approach authorizes dynamic camera position adjustment, ensuring acquisition of the most informative viewpoint.

Ultimately, we introduce a comprehensive task flow for robotic perception and manipulation of articulated objects, as shown in Fig.~\ref{Graphical summary}. The process starts with the robot selecting an object part, then inputting corresponding RGB images and point cloud data into the perception network to determine the part's movability and identify joint parameters. If deemed movable and the perception score exceeds a threshold, the robot plans a manipulation sequence based on the identified joint parameters as well as the operation command. Conversely, when suboptimal viewing angles are detected, the robot adjusts its viewpoint for further assessment, ensuring reliable, precise interactions with the articulated objects.

To rigorously test our model, experiments were conducted on Sapien simulation platform~\cite{Xiang_2020_SAPIEN} using PartNet-Mobility~\cite{Mo_2019_CVPR}. Our method showed significant advancements, evident in notable performance improvements over existing benchmarks. Key contributions are summarized as follows.

\begin{itemize}
	\item We developed a multimodal feature fusion technique that significantly enhances point feature representation by integrating multi-scale image details, thus substantially improving representational capability.
	\item We developed an active sensing technique grounded in reinforcement learning, empowering the robot to autonomously optimize its camera position to capture the most informative viewpoint.
	\item MARS achieved state-of-the-art in joint parameter estimation for diverse articulated objects. It effectively perceives both revolute and prismatic joints using a single network, validated in real-world applications.
\end{itemize}

\section{Related Works}
\paragraph{Articulated Object Characterization.}
To enhance robotic capabilities in perceiving and manipulating articulated objects, a wealth of simulators~\cite{todorov2012mujoco,Xiang_2020_SAPIEN,szot2021habitat} and open datasets~\cite{chang2015shapenet,Mo_2019_CVPR,wang2019shape2motion,geng2023gapartnet} have emerged as essential resources. These tools have facilitated advancements in 3D reconstruction~\cite{li2020category,bozic2021neural,yang2021lasr}, joint parameter estimation~\cite{jain2021screwnet,shi2021self,zeng2021visual,jiang2022ditto,chu2023command}, and the prediction of interactive positions and trajectories~\cite{mo2021where2act,wu2021vat,wang2022adaafford} for articulated objects. 

In joint parameter estimation, prior work~\cite{yi2018deep,abbatematteo2019learning,shi2021self,jain2021screwnet,siarohin2021motion,jiang2022ditto} has leveraged multi-view observations from ongoing monitoring as visual inputs, capitalizing on the rich visual cues provided by the changing joint states. However, this method complicates data collection, limiting its practicality in real-world robotics. Furthermore, many studies~\cite{yan2020rpm,jiang2022ditto,chu2023command} rely exclusively on point cloud data, neglecting the valuable features available in RGB imagery that can enhance joint parameter estimation. Our methodology focuses on employing a single RGB image along with point cloud data to achieve accurate perception of joint parameters.

\paragraph{Multimodal Feature Fusion.}
In the realm of multimodal feature fusion, especially in combining RGB and point cloud data, current research predominantly focuses on 3D object detection~\cite{ku2018joint,sindagi2019mvx,zhu2021cross,piergiovanni20214d,wu2022sparse,zhang2022maff}. However, due to differences in input data, these fusion methods often struggle to be directly applicable and effective in joint analysis. A notable example similar to our approach is EPNet~\cite{huang2020epnet}, which enhances point cloud features by integrating image features into the point cloud domain, utilizing both local and global contexts of images. In contrast, our method employs a competitive mechanism to dynamically prioritize the significance of image features across various scales.

\paragraph{Active Sensing.}
In practical applications, the challenging task of obtaining optimal imaging angles for robots~\cite{7989164,9729641} underscores the need for active sensing. This technique entails dynamically adjusting sensor positions or the manipulation environment to enhance data acquisition, proving essential for certain tasks~\cite{8601345,9561650,9636414}, especially in scenarios where initial views are insufficient due to partial occlusions or limited visibility of articulated objects. In current studies, researchers commonly assume that high-quality images and point clouds can be obtained from an ideal viewpoint~\cite{jain2021screwnet,yang2021lasr,jiang2022ditto,chu2023command}, but such assumptions frequently fail to align with the reality. Our approach successfully adapts to complex real-world application scenarios by incorporating active sensing techniques.

\section{Method}
\begin{figure*}[!ht]
	\centering
	\includegraphics[width=6.9in,keepaspectratio]{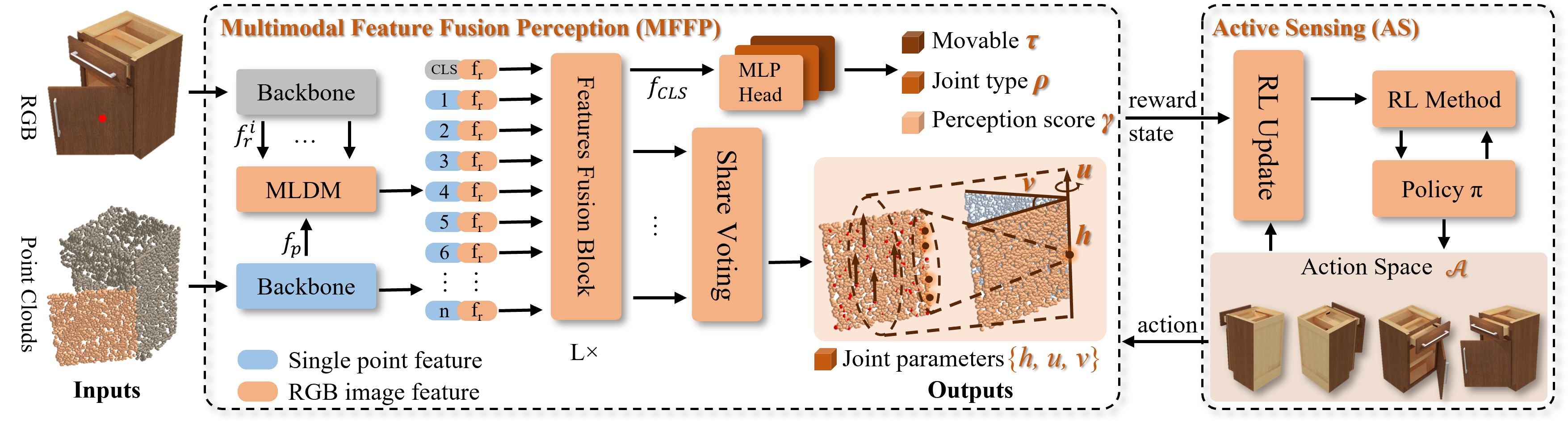}
	\caption{MARS Framework with MFFP and AS components. MFFP integrates RGB and point cloud data, utilizing MLDM for adaptive RGB feature scaling (see Fig.~\ref{MLDM}) and a Fusion Block for combining features, aiding in joint parameters prediction. AS adjusts viewpoints under suboptimal conditions, enhancing perception accuracy in real-world scenarios.}
	\label{framework}
\end{figure*}

We introduce MARS, a framework designed for estimating joint parameters in articulated objects. As depicted in Fig.~\ref{framework}, MARS consists of two primary components: Multimodal Feature Fusion Perception (MFFP) and Active Sensing (AS).

The MFFP component of MARS utilizes ResNet18~\cite{he2016deep} and PointNet++~\cite{qi2017pointnet++} as backbone networks to efficiently extract features from the input RGB image and point cloud data. The Multi-Layer Dueling Module (MLDM) is designed to strategically extract and weigh image features across different scales. Following this, image and point cloud features are merged at the feature level via a specialized fusion module. A decoding module then processes these integrated features to produce outputs for joint parameters and a perception score, reflecting the efficacy of the current viewpoint.

The AS module operates based on the perception score and a set threshold. If the score is below this threshold, it triggers a viewpoint change. A new observation position is determined from the action space, followed by a repeated perception process. This approach enables real-time adjustments for optimal data acquisition and improved joint parameter estimation in complex or obstructed scenarios.

\subsection{MLDM}
\begin{figure}[!t]
	\centering
	\includegraphics[width=2.85in,keepaspectratio]{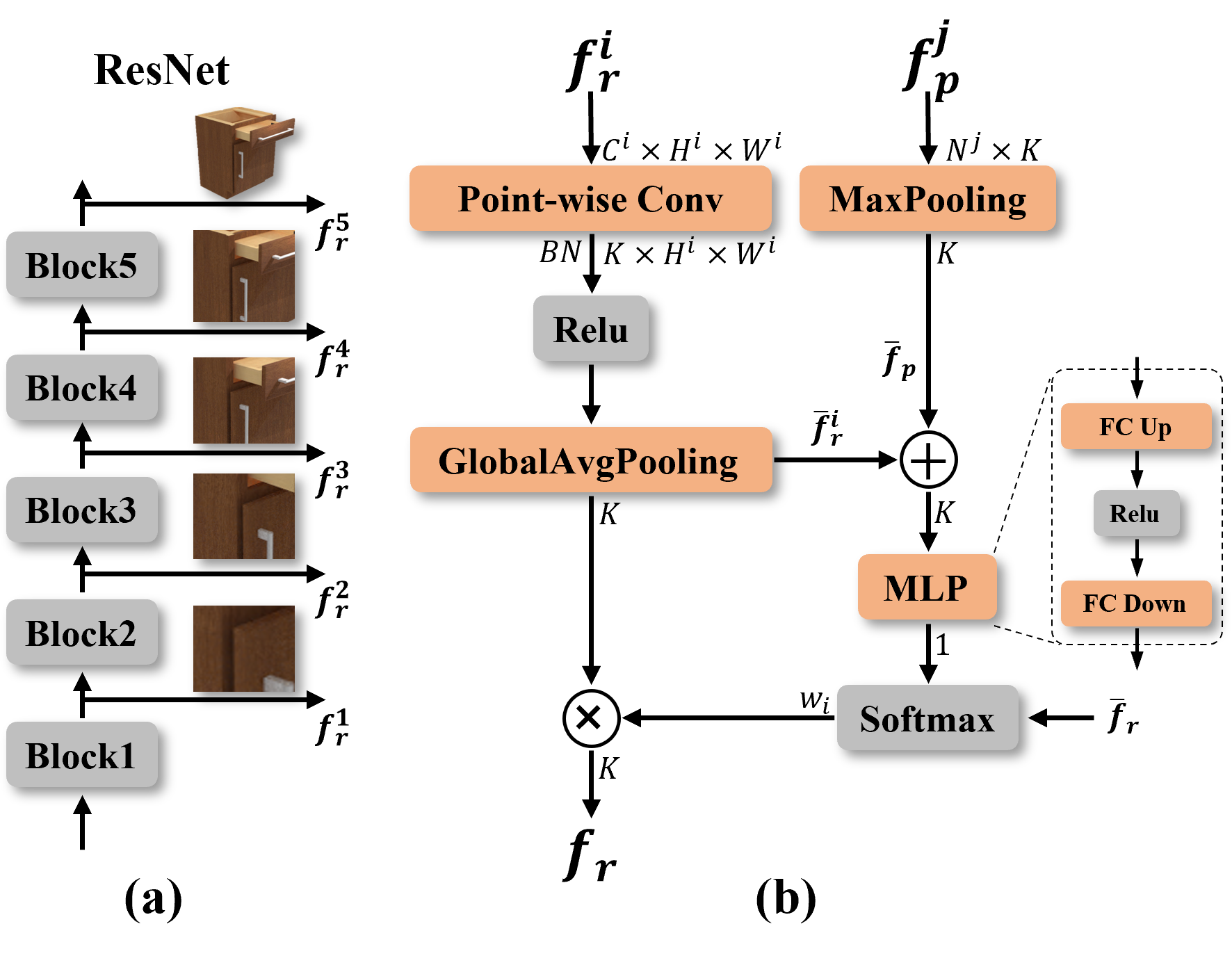}
	\caption{MLDM Architecture for RGB Image Feature Aggregation. (a) Image feature maps $f^i_{r}$ are extracted at multiple scales from ResNet blocks. (b) These feature maps are first combined with point cloud feature $f^j_p$, upon which adaptive weights $w_i$ are computed to form the final weighted image feature representation $f_r$.}
	\label{MLDM}
\end{figure}

Employing ResNet, we capturing a range of articulated object RGB features from local to global, as depicted in Fig.~\ref{MLDM}a. For each feature map $f^i_r \in \mathbb{R}^{C^i \times H^i \times W^i}$ with $C^i$ channels and spatial dimensions $H^i \times W^i$, and for point cloud features $f^j_p \in \mathbb{R}^{N^j \times K}$ comprising $N^j$ points each of $K$ dimensions, we perform point-wise convolution and subsequent pooling operations to aggregate the features (as shown in Fig.~\ref{MLDM}b). The aggregated image feature $\bar{f^i_r} \in \mathbb{R}^K$ is computed as follows:
\begin{equation}
	\bar{f^i_r}=\mathbf{g}\left(\delta\left(\mathcal{B}\left(\operatorname{PWConv}(f^i_r)\right)\right)\right),
\end{equation}
where $\mathbf{g}$ is the global average pooling, $\delta$ is the ReLU function~\cite{nair2010rectified}, and $\mathcal{B}$ represents BN~\cite{ioffe2015batch}, with the kernel size for the PWConv being $K \times 1 \times 1$. For point cloud feature aggregation, the max pooling operation $\mathbf{m}$ is utilized:
\begin{equation}
	\bar{f_p}=\mathbf{m}\left(f^j_{p}\right).
\end{equation}

We combine the aggregated features $\bar{f^i_{r}}$ and $\bar{f_{p}}$, and use a Multi-Layer Perceptron (MLP) to determine the weights for the final RGB image feature $f_{r} \in \mathbb{R}^K$. The computation is given by:
\begin{equation}
	f_{r} = \sum_{i=1}^n \left(\bar{f^i_{r}} \otimes \sigma \left(\theta_{w}\left(\bar{f^i_{r}} + \bar{f_{p}}\right)\right)\right),
\end{equation}
where $\sigma$ represents the softmax function, and $\theta_{w}$ denotes the learnable parameters within the MLP. Comprising two linear layers with a ReLU activation between them, the MLP scales up and then reduces the dimensions of the fused feature to compute the weight $w_i \in \mathbb{R}^1$. The operation $\otimes$ indicates element-wise multiplication. The weights from all blocks are then combined to form the final weight distribution $w$.

\subsection{Feature Fusion Block}
This component leverages transformer encoders~\cite{vaswani2017attention}, intentionally excluding positional embedding due to the intrinsic spatial information of the RGB and point cloud features. The input tokens $t^j = f^j_p \oplus f_r \in \mathbb{R} ^ {N^j \times 2K}$ for the fusion module are formed by concatenating image feature to each point feature. A CLS token $t^{\text{CLS}}$ is also incorporated to encapsulate global features, which is pivotal for the model's generalization. The Feature Fusion Block outputs a global feature $f^{\text{CLS}} \in \mathbb{R}^{2K}$ and local features $\widetilde{f}^j_p \in \mathbb{R}^{N^j \times 2K}$ for each point, expressed as:
\begin{equation}
	\{f^{\text{CLS}}, \widetilde{f}^j_p\} = \mathbf{F}_L(\mathbf{F}_{L-1}(\ldots \mathbf{F}_1(t^{\text{CLS}}, t^j) \ldots)),
\end{equation}
where $\mathbf{F}_l$ is the $l$-th layer of the fusion block.

\subsection{Articulation Decoders}
We utilize the global feature $f_{\text{CLS}}$ to determine the movability of the chosen rigid part, represented as a binary variable $\tau \in {0, 1}$, where $\tau = 0$ signifies an immovable part and $\tau = 1$ a movable part. An MLP head dedicated to this task decodes this parameter:
\begin{equation}
	\tau = \theta_{mov} \left(f_{\text{CLS}}\right)
\end{equation}

\paragraph{Joint Parameters.}
Upon determining the movability of the target part, we assume a fully closed position as the initial state to estimate the joint parameters. These include the joint type $\rho \in {0, 1}$, where $\rho = 0$ indicates a revolute joint and $\rho = 1$ a prismatic joint, the joint position $h \in \mathbb{R}^3$, orientation $u \in \mathbb{R}^3$, and the current state $v \in \mathbb{R}$. Unlike previous models that omit position predictions for prismatic joints~\cite{jiang2022ditto,chu2023command}, our framework accommodates both revolute and prismatic joints, identifying the centroid of the movable part as the position for the prismatic joint. The joint type $\rho$ is deduced from $f_{\text{CLS}}$ using an MLP head. For the joint parameters, a shared voting module leverages each point $p^j \in \mathbb{R}^{N^j \times 3}$ and its corresponding feature $\widetilde{f}^j_p \in \mathbb{R}^{N^j \times 2K}$ for point-wise voting to infer $h^j$, $u^j$, and $v^j$. The final joint parameters are then the mean of these votes:
\begin{equation}
	\begin{aligned}
		& \rho = \theta_{type} \left(f_{\text{CLS}}\right), \\
		& \{h^j, u^j, v^j\} = \theta_{para} \big(p^j, \widetilde{f}^j_p\big), \\
		& \{h, u, v\} = \frac{1}{N} \{\sum_{j=1}^{N} h^j, \sum_{j=1}^N u^j, \sum_{j=1}^N v^j\} .
	\end{aligned}
\end{equation}

\paragraph{Perception Score.}
To assess the quality of perception results, we use an additional MLP head with $f_{\text{CLS}}$ as the input. This module predicts the likelihood of successful perception $\gamma \in (0,1)$ in the following manner:
\begin{equation}
	\gamma = \zeta(\theta_{score}(f_{\text{CLS}})),
\end{equation}
where $\zeta$ is the sigmoid activation function, ensuring $\gamma$ falls between 0 and 1. A success threshold of 0.5 is set for binary decision-making within the module.

\paragraph{Loss Functions.}
To supervise the predictions of movability and joint type, binary cross-entropy loss functions $\mathcal{L}_{mov}$ and $\mathcal{L}_{type}$ are utilized. For joint state prediction, an L1 norm loss function $\mathcal{L}_{state}$ is employed. As the perception score prediction has been transformed into a binary classification task for assessing viewpoint optimality, it is supervised using binary cross-entropy loss $\mathcal{L}_{score}$. To penalize the discrepancy in orientation between the estimated joint and the ground truth $\hat{u}$, which is a unit vector, the loss $\mathcal{L}_{ori}$ is defined as:
\begin{equation}
	\mathcal{L}_{ori} = \frac{1}{N} \sum_{j=1}^N \text{arccos}\left(\hat{u} \cdot u^j\right),
\end{equation}
The loss $\mathcal{L}_{pos}$ penalizes the distance between the estimated projection point $h^j$ and the actual joint axis $\hat{u}$:
\begin{equation}
	\mathcal{L}_{pos} = \frac{1}{N} \sum_{j=1}^N || (h^j - \hat{h}) \times \hat{u} ||.
\end{equation}

\paragraph{Training Steps.}
In the initial training phase, we pre-train the model parameters using the movability loss $\mathcal{L}_{mov}$ along with the corresponding decoding head. This step is crucial for ensuring efficient learning of movability prediction. For joint parameter prediction, the model is trained to minimize the differences between the predicted and the actual ground truth values. The cumulative loss for this training phase is given by:
\begin{equation}
	\mathcal{L}_{para} = \mathcal{L}_{type} + \mathcal{L}_{ori} + \mathcal{L}_{pos} + \mathcal{L}_{state}
\end{equation}
In the final training stage, parameters except for the perception scoring decoder are fixed, with the training focused on optimizing the scoring decoder using the loss $\mathcal{L}_{score}$.

\begin{figure}[!t]
	\centering
	\includegraphics[width=3.3in,keepaspectratio]{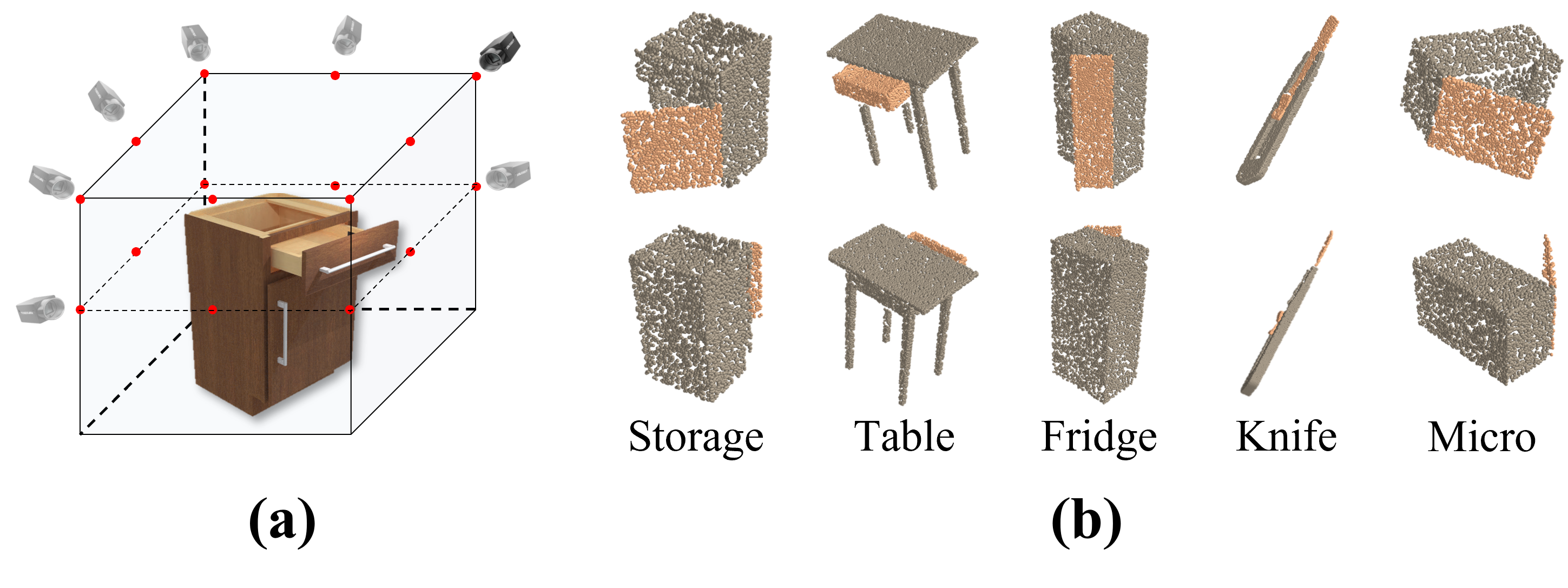}
	\caption{(a) Sixteen discrete positions constitute the discrete action space for viewpoint selection. (b) Comparative analysis of point cloud quality obtained from different viewpoints.}
	\label{action and compare}
\end{figure}

\subsection{The RL Policy For Active Sensing}
Training of a conditional RL policy, using a DQN approach~\cite{mnih2015human}, follows the completion of perception module training. This policy, aimed at active viewpoint optimization, is trained within a simulated environment. Success in a training iteration occurs when the perception score exceeds 0.5; over 5 action steps indicate failure. Details of the RL policy are provided in the following section.
\paragraph{State Space.}
In the state space, a pre-trained perception network processes an RGB image and point cloud from the current viewpoint, yielding the global feature $f_{\text{CLS}}$. Additionally, the position of camera is represented as a one-hot encoded vector $x \in \mathbb{R}^{16}$, indicating one of 16 distinct positions. The state input for the RL algorithm is formed by the combination of $f_{\text{CLS}}$ and this one-hot encoded position $x$.
\paragraph{Action Space.}
	We established a discrete action space $\mathcal{A} \in \{0, 1\}^{16}$ with 16 viewpoints around the object. This action space aligns with the coordinate framework of the object, as depicted in Fig.~\ref{action and compare}a. To better simulate a real environment, some locations will be randomly set as unreachable.
\paragraph{Reward Design.}
The per-step rewards $r_{step}$ are determined by two primary criteria: perception score and point cloud quantity variation. To balance these aspects, the step reward is defined as:
\begin{equation}
	r_{step} = \lambda_s r_{score} + \lambda_n r_{num},
\end{equation}
where $\lambda_s$ and $\lambda_n$ are weight coefficients that adjust the relative importance of $r_{score}$ and $r_{num}$ in the reward. Here, $r_{score} = s - s'$ represents the change in perception scores before and after an action, and the point cloud variation reward is given by:
\begin{equation}
	r_{num} = \frac{n - n'}{n'} - 1,
\end{equation}
where n denotes the number of point clouds. Additionally, at the end of a round, a positive reward of $+10$ is assigned for successful task completion. Conversely, a negative reward of $-10$ is incurred if the task fails due to exceeding the action step limit.

\begin{table*}[t!]
	\centering
	\footnotesize
	\setlength{\tabcolsep}{1.4pt}
	\begin{tabular}{l|cccccccccc|cccc|c}
		\toprule
		\multirow{2}{*}{Methods} & \multicolumn{10}{c|}{Revolute} & \multicolumn{4}{c|}{Prismatic} & \multirow{2}{*}{Avg} \\
		&Laptop&Chair&Pliers&Safe&Eyeglass&Fridge&Scissor&Door&Micro&Oven&Table&Storage&Window&Knife&\\
		\midrule
		\multicolumn{16}{c}{Errors of Joint Position}\\
		\midrule
		RPM-Net     & 0.12 & 0.10 & 0.13 & 0.16 & 0.18 & 0.14 & 0.17 & 0.25 & 0.16 & 0.21 & - & - & - & - & 0.16/- \\
		ANCSH       & 0.08 & - & - & - & 0.06 & - & - & - & 0.06 & 0.12  & - & - & - & - &  0.08/- \\
		Ditto       & 0.03 & \textbf{0.02} & \textbf{0.03} & 0.06 & 0.04 & 0.06 & 0.04 & 0.06 & \textbf{0.03} & 0.03 & - & - & - & - & 0.04/- \\
		Cart        & 0.03 & 0.03 & \textbf{0.03} & 0.06 & \textbf{0.03} & \textbf{0.04} & 0.03 & 0.06 & 0.05 & 0.03 & - & - & - & - & 0.04/- \\
		EPNet*       & 0.03 & 0.03 & 0.04 & \textbf{0.05} & 0.05 & 0.05 & 0.03 & 0.07 & \textbf{0.03} & 0.04 & 0.15 & 0.25 & 0.04 & 0.03 & 0.04/0.12 \\
		\midrule
		MFFP(Ours)  & \textbf{0.02} & \textbf{0.02} & \textbf{0.03} & \textbf{0.05} & \textbf{0.03} & 0.05 & \textbf{0.02} & \textbf{0.05} & \textbf{0.03} & \textbf{0.02} &  \textbf{0.13} & \textbf{0.21} & \textbf{0.02} & \textbf{0.01} & \textbf{0.03/0.09} \\
		w/o MLDM        & 0.03 & 0.03 & \textbf{0.03} & 0.06 & 0.04 & 0.05 & \textbf{0.02} & 0.06 & \textbf{0.03} & 0.04 & 0.15 & 0.22 & 0.04 & 0.03 & 0.04/0.11 \\
		\midrule
		\multicolumn{16}{c}{Errors of Joint Orientation}\\
		\midrule
		RPM-Net     & 10.73$^\circ$ & 11.69$^\circ$ & 7.83$^\circ$ & 12.64$^\circ$ & 20.52$^\circ$ & 6.27$^\circ$ & 4.85$^\circ$ & 16.38$^\circ$ & 7.42$^\circ$ & 5.28$^\circ$ & 17.37$^\circ$ & 9.63$^\circ$ & 7.84$^\circ$ & 5.92$^\circ$ & 10.36$^\circ$/10.19$^\circ$ \\
		ANCSH       & 3.92$^\circ$  & - & - & - & 5.87$^\circ$  & - & - & - &3.64$^\circ$ & 3.15$^\circ$ & - & 7.74$^\circ$ & - & - & 4.15$^\circ$/7.74$^\circ$  \\
		Ditto       & 1.52$^\circ$ & 2.54$^\circ$ & 1.42$^\circ$ & 2.98$^\circ$ & 1.87$^\circ$ & 2.21$^\circ$ & 1.33$^\circ$ & 1.96$^\circ$ & 1.30$^\circ$ & 1.70$^\circ$ & 1.83$^\circ$ & 4.74$^\circ$ & 2.73$^\circ$ & 1.76$^\circ$ & 1.88$^\circ$/2.77$^\circ$ \\
		Cart        & 1.45$^\circ$ & 2.78$^\circ$ & 1.23$^\circ$ & 2.04$^\circ$ & 1.82$^\circ$ & \textbf{2.03$^\circ$} & 1.26$^\circ$ & 2.48$^\circ$ & 1.26$^\circ$ & 1.78$^\circ$ & 1.78$^\circ$ & 4.03$^\circ$ & 2.56$^\circ$ & 1.69$^\circ$ & 1.81$^\circ$/2.52$^\circ$ \\
		EPNet*       & 0.94$^\circ$ & 1.69$^\circ$ & 1.15$^\circ$ & 1.53$^\circ$ & 1.27$^\circ$ & 4.38$^\circ$ & 1.16$^\circ$ & 1.79$^\circ$ & 0.83$^\circ$ & 1.02$^\circ$ & 2.37$^\circ$ & 4.14$^\circ$ & 0.18$^\circ$ & 0.16$^\circ$ & 1.58$^\circ$/1.71$^\circ$ \\
		\midrule
		MFFP(Ours) &\textbf{0.08$^\circ$} &\textbf{0.44$^\circ$} &\textbf{0.06$^\circ$} &\textbf{0.24$^\circ$}  &\textbf{0.39$^\circ$}  &3.36$^\circ$ & \textbf{0.05$^\circ$} &\textbf{1.51$^\circ$}  &\textbf{0.28$^\circ$} &\textbf{0.10$^\circ$}  &\textbf{0.42$^\circ$} &\textbf{2.96$^\circ$} &\textbf{0.04$^\circ$} &\textbf {0.04$^\circ$} & \textbf{0.65$^\circ$/0.87$^\circ$} \\
		w/o MLDM        & 0.56$^\circ$ & 1.78$^\circ$ & 0.95$^\circ$ & 1.24$^\circ$ & 1.72$^\circ$ & 4.16$^\circ$ & 0.68$^\circ$ & 3.23$^\circ$ & 2.24$^\circ$ & 0.87$^\circ$ & 1.08$^\circ$ & 4.27$^\circ$ & 0.25$^\circ$& 0.56$^\circ$ & 1.74$^\circ$/1.54$^\circ$ \\
		\midrule
		\multicolumn{16}{c}{Errors of Joint State}\\
		\midrule
		RPM-Net   & 15.38$^\circ$ & 17.99$^\circ$ & 14.66$^\circ$ & 9.96$^\circ$ & 18.85$^\circ$ & 16.45$^\circ$ & 8.24$^\circ$ & 25.78$^\circ$ & 19.91$^\circ$ & 14.24$^\circ$ & 0.27 & 0.31 & 0.38 & 0.11 & 16.15$^\circ$/0.27 \\
		ANCSH       & 10.15$^\circ$ & - & - & - & 15.74$^\circ$ & - & - & - & 14.72$^\circ$ & 9.03$^\circ$ & - & 0.24 & - & - & 12.41$^\circ$/0.24 \\
		Ditto       & 3.98$^\circ$ & 4.98$^\circ$ & 3.89$^\circ$ & 4.18$^\circ$ & 7.73$^\circ$ & 6.14$^\circ$ & 3.05$^\circ$ & 13.53$^\circ$ & 6.80$^\circ$ & 3.86$^\circ$ & 0.05 & \textbf{0.07} & 0.16 & 0.02 & 5.81$^\circ$/0.08 \\
		Cart        & 3.64$^\circ$ & 5.53$^\circ$ & 3.14$^\circ$ & 4.20$^\circ$ &  6.43$^\circ$  & 5.83$^\circ$ & 2.79$^\circ$ & \textbf{11.07$^\circ$} & \textbf{3.74$^\circ$} & 3.18$^\circ$ & \textbf{0.04} & 0.08 & 0.17 & 0.02 & 4.96$^\circ$/0.08 \\
		EPNet*       & 1.96$^\circ$ & 3.79$^\circ$ & 2.37$^\circ$ & 3.53$^\circ$ & 3.79$^\circ$ & \textbf{5.25$^\circ$} & 2.53$^\circ$ & 13.49$^\circ$ & 5.53$^\circ$ & 2.64$^\circ$ & 0.09 & 0.08 & 0.18 & 0.02 & 4.49$^\circ$/0.09 \\
		\midrule
		MFFP(Ours)  & \textbf{1.21$^\circ$} & \textbf{3.20$^\circ$} & \textbf{1.98$^\circ$} & \textbf{2.38$^\circ$} & \textbf{3.13$^\circ$} & 5.30$^\circ$ & \textbf{1.94$^\circ$} & 12.03$^\circ$ & 4.88$^\circ$ & \textbf{1.34$^\circ$} & 0.05 & \textbf{0.07} & \textbf{0.13} & \textbf{0.01} & \textbf{3.74$^\circ$/0.07} \\
		w/o MLDM        & 1.79$^\circ$ & 3.78$^\circ$ & 2.05$^\circ$ & 2.94$^\circ$ & 4.24$^\circ$ & 5.80$^\circ$ & 2.39$^\circ$ & 15.48$^\circ$ & 5.51$^\circ$ & 2.28$^\circ$ & 0.06 & 0.08 & 0.20 & 0.02 & 4.63$^\circ$/0.09 \\
		\bottomrule
	\end{tabular}
	\caption{Quantitative evaluation of joint parameters estimation across 14 categories. The best results are in bold, with our method demonstrating superior performance in most categories.}
	\label{Quantitative evaluation}
\end{table*}

\subsection{Command-Based Point Cloud Manipulation}
We use perceived joint parameters for point cloud manipulations based on operational commands $\mathcal{C}$. The system modifies the joint state in response to $\mathcal{C}$, involving adjustments such as angle and position changes. Specifically, $\mathcal{C}$ specifies the target state for the joint, and we calculate the difference between this target and the current perceived joint state as the operational compensation $\Delta v = \mathcal{C} - v$. Based on $\Delta v$, joint orientation $u$, and joint position $h$, we compute a rotation-translation matrix $\mathcal{M}$ to represent the current operation. The matrix is calculated as follows:
\begin{equation}
	\mathcal{M} = \begin{bmatrix}
		R(\Delta v, u) & T(\Delta v, h) \\
		0 & 1
	\end{bmatrix},
\end{equation}
where $R(\Delta v, u)$, the rotation matrix, is derived from $\Delta v$ and joint orientation $u$. $T(\Delta v, h)$, the translation vector, is calculated from $\Delta v$ and joint position $h$.

\section{Experimental Evaluation}
We evaluated perception capabilities of MARS for articulated objects. Quantitative assessment across various object categories confirmed accuracy of the MFFP module in estimating joint parameters. Integration of active sensing for viewpoint optimization significantly enhanced algorithm performance. Visual demonstrations of command-based point cloud manipulation and qualitative showcasing of method effectiveness on real-world objects were also conducted.

\begin{figure*}[!t]
	\centering
	\includegraphics[width=6.9in,keepaspectratio]{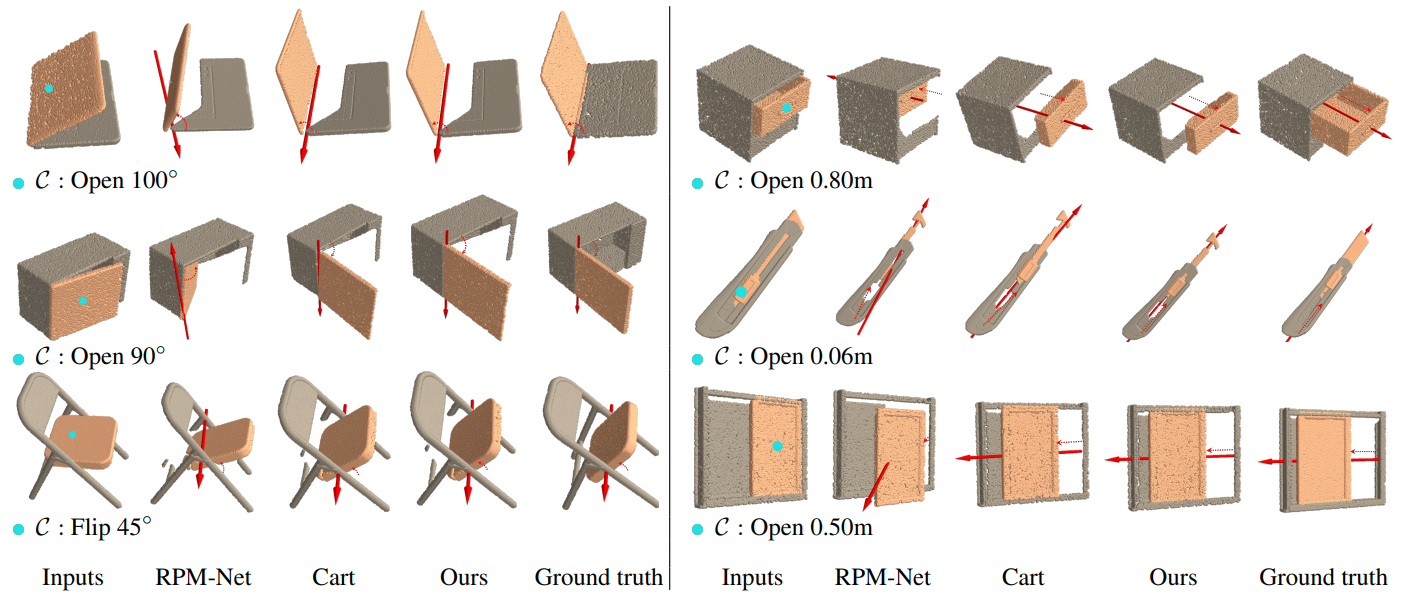}
	\caption{Comparison of point cloud-level manipulation visualizations, where blue dots represent selected parts and $\mathcal{C}$ denotes the current manipulation command.}
	\label{visualization}
\end{figure*}

\subsection{Experimental Setup}
\paragraph{Datasets.}
For evaluation, we utilized the SAPIEN simulator~\cite{Xiang_2020_SAPIEN} and PartNet-Mobility dataset~\cite{Mo_2019_CVPR}, selecting 14 common articulated objects (10 with revolute and 4 with prismatic joints). In the simulator, these objects, with randomized joint states and camera positions, generated various viewpoint samples (see Fig.~\ref{action and compare}b). Post movability prediction module training, immovable parts data was removed, resulting in 10K training, 1K testing, and 1K validation samples for each category for perception network training. Image data was captured at 600 $\times$ 600 resolution using an RGB-D camera.
\paragraph{Baselines.}
We benchmarked our method against six approaches. RPM-Net~\cite{yan2020rpm} employs recurrent neural networks for predicting object motion from point clouds. ANCSH~\cite{li2020category} focuses on joint parameter estimation in canonical object space. Ditto~\cite{jiang2022ditto} utilizes multi-view for articulated object understanding. Cart~\cite{chu2023command}, the current state-of-the-art (SOTA), specializes in joint parameter estimation. To facilitate prediction of the current joint state by Cart, a 'Closed' command was issued in each evaluation. EPNet~\cite{huang2020epnet}, a multimodal fusion approach for 3D object detection, was adapted by replacing our MLDM module with its LiDAR-guided Image Fusion (LI-Fusion) module, enabling EPNet* to estimate joint parameters. Additionally, we included an ablated version of MARS lacking MLDM for comparison. Lastly, we proportionally increased poor viewpoints data for all methods to ensure a fair comparison.
\paragraph{Evaluation Metrics.}
Our investigation primarily focuses on the estimation errors of articulated object joint parameters. Specifically, we measure errors in estimating joint orientation, joint position, and the current joint state of the selected part relative to its fully closed initial state.
\paragraph{RL Environment.}
To train and validate active sensing strategies, a reinforcement learning environment was constructed. In each training round, an object is randomly imported into the simulation environment, and a camera position is initialized within the action space with a step size limit of 5. For testing purposes, the step size is set to 1. If the initial sensing score falls below a predefined threshold, the robot selects a new observation position based on the action strategy.

\subsection{Main Results.}
\paragraph{Comparison of joint parameters estimation.}
Table~\ref{Quantitative evaluation} shows the joint parameter estimation results from our quantitative evaluation, where our method surpasses the SOTA in most categories and closely matches it in the rest. This superior performance can be attributed to our robust feature representation, a result of the multimodal fusion approach that effectively utilizes the rich information in RGB images to enhance point cloud features. Additionally, in comparison to EPNet*, our MLDM exhibits enhanced multi-scale feature extraction capabilities, significantly improving articulated object perception. It's important to note that joint state estimation is the most challenging, with the largest errors. Our method leads in performance, yet encounters a 3.74° error in revolute joints and 0.07m in prismatic joints. The large error mainly arises from poor observation perspectives in the acquired data, as depicted in Figure 4b. Practically, robots often encounter such angles, highlighting the importance of active sensing to adjust the viewpoint.
\paragraph{Ablation Studies.}
Table~\ref{Quantitative evaluation} demonstrates evaluation of the MLDM module impact on performance. Across all categories, the complete version outperformed the ablated version in joint parameter estimation. Notably, even the ablated version marginally surpassed other approaches, highlighting the effectiveness of multimodal fusion with RGB data combined with limited point cloud input.

\begin{table}[!h]
	\centering
	\footnotesize
	\setlength{\tabcolsep}{4pt}
	\begin{tabular}{llcccc}
		\toprule
		\multicolumn{2}{c}{Class} &Type acc &Position &Orientation &State\\
		\midrule
		\multirow{2}{*}{Revolute}  &Micro & 100\% & 0.03 & 0.35$^\circ$ & 4.05$^\circ$ \\
		&Oven & 100\% & 0.03 & 0.27$^\circ$ & 1.37$^\circ$ \\
		\midrule						   
		\multirow{2}{*}{Prismatic} &Table & 100\% & 0.14 & 2.63$^\circ$ & 0.06 \\
		&Storage & 100\% & 0.25 & 4.57$^\circ$ & 0.07 \\
		\bottomrule
	\end{tabular}
	\caption{Results from harmonized training on mixed joint types}
	\label{mixed joint types}
\end{table}

\paragraph{Harmonized Training for Mixed Joint Types.}
To address the need for separate models for revolute and prismatic joints in current approaches, joint parameter representation was standardized by defining prismatic joint positions at centroids of movable parts. As demonstrated in Table~\ref{mixed joint types}, with balanced training samples for both joint types, 100\% accuracy was achieved in joint type prediction. Performance slightly declined compared to models trained on each joint type separately, yet notable results were still achieved in quantitative joint parameter evaluation.

\begin{figure*}[!ht]
	\centering
	\includegraphics[width=7in,keepaspectratio]{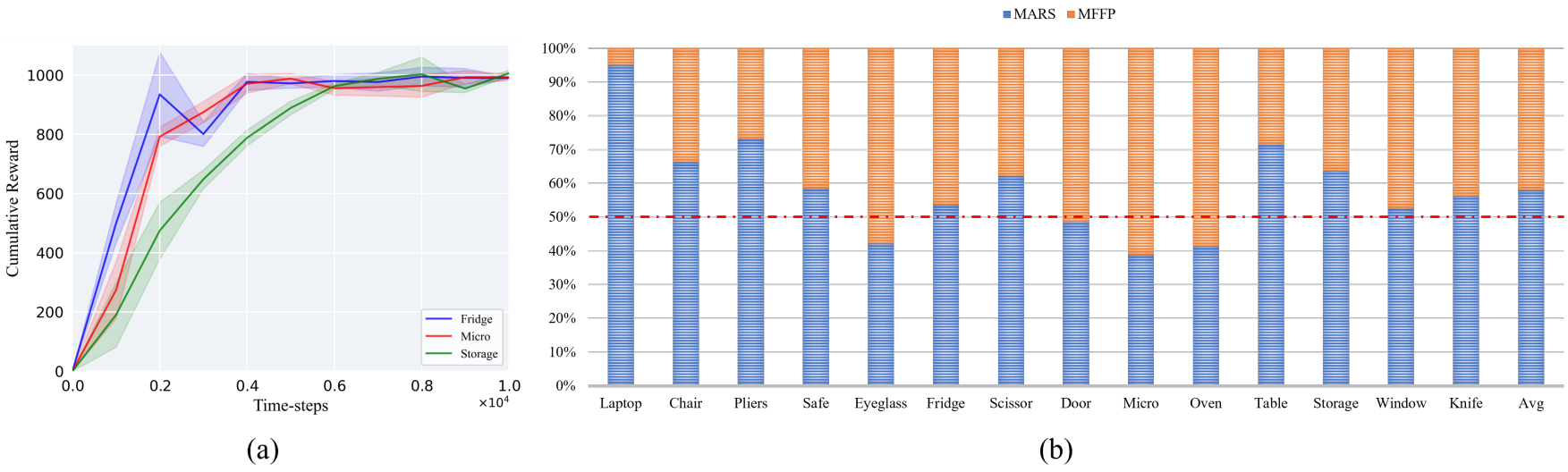}
	\caption{(a)Reward trends across there categories;(b)Bar chart showing joint state error reduction using active sensing in MARS versus MFFP, with MFFP normalized to 100\% and the red line at 50\%.}
	\label{RLAS}
\end{figure*}

\begin{figure}[!ht]
	\centering
	\includegraphics[width=3.2in,keepaspectratio]{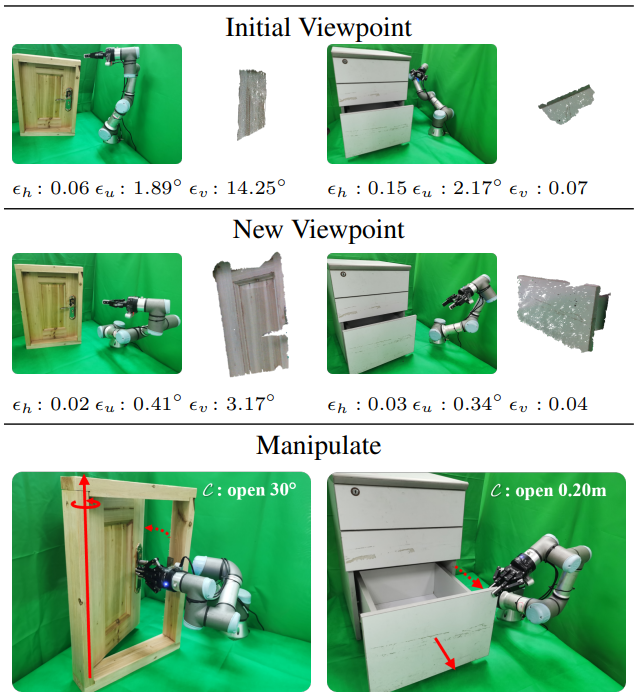}
	\caption{Real world Experiments.}
	\label{Real}
\end{figure}

\paragraph{Visualization results.}
We visualized and evaluated point cloud level manipulations by first specifying the desired joint state as a command, such as 'open $100^{\circ}$'. Then, we executed the corresponding manipulation on the point cloud of the selected part using the rotation-translation matrix computed as per Equation (13). Fig.~\ref{visualization} compares the performance of RPM-Net~\cite{yan2020rpm}, Cart~\cite{chu2023command}, and our method against the ground truth, with blue dots indicating the selected parts. Overall, our method more accurately reflects the real situation post-manipulation. In comparison, Cart tends to have larger state estimation errors, resulting in poorer maneuvering performance. RPM-Net often incorrectly predicts the joint direction, leading to significant discrepancies with actual ground conditions.

\subsection{Enhancing Results through Active Sensing}
We compared the performance of our method with and without the active sensing module. To expedite testing, we pre-collected samples from each articulated object instance at 16 camera positions after randomizing joint states. For each instance, the initial camera viewpoint provided a perception result to establish MFFP performance. A single active sensing adjustment informed by the perception score followed, marking the outcome of this camera position change as MARS performance. Fig.~\ref{RLAS}a charts the learning curves for active sensing in three categories, with cumulative rewards increasing and eventually converging over time, a pattern echoed across other categories not depicted. Fig.~\ref{RLAS}b presents a performance comparison of the two models, where the estimation error for the most challenging joint state in each articulated object class is normalized to 100\% for MFFP. Bar graphs subsequently illustrate the proportional error reduction achieved by MARS. Overall, MARS achieved an average error reduction of approximately 50\% across all categories, particularly notable in categories of articulated objects with larger sizes and inherent self-occlusion issues. This analysis underscores the efficacy of active sensing in enhancing the accuracy of our method, demonstrating its vital role in complex joint parameter estimation.

\subsection{Real-world Experiments}
To validate the generalizability of our method in real-world settings, we selected two articulated objects: a door and a table with drawers. As depicted in Fig.~\ref{Real}, we used an Intel RealSense RGB-D camera on a mobile robot to capture RGB and point cloud data of these objects. Segmentation of the rigid parts was achieved using 3D U-Net~\cite{choy20194d}. The perception scoring module assessed the input viewpoint quality. If the score fell below the threshold, active sensing guided the robot to a new position ID for additional data acquisition. Once the perception score exceeded the threshold, the robot formulated its action plan based on the perception results and specified commands. Interaction positions between the robot and the object were manually assigned, leading to the successful manipulation of the target object.

\section{Conclusion}
In this paper, we introduced MARS, a multimodal framework specifically designed for accurately sensing joint parameters of articulated objects. Central to this framework is the MLDM, an innovative approach for adaptive multiscale feature fusion that significantly enhances image feature representation. MARS utilizes a transformer encoder, devoid of positional embedding, to effectively integrate RGB features with point cloud data.The significant advancement in this work is the reinforcement learning-based active perception strategy, empowering robots to autonomously seek new perspectives and substantially improve practical applicability in response to inadequate perception. Future research aims to enhance MARS by seeking more powerful point cloud representation capabilities and improving algorithmic generalization to cover a broader range of articulated objects.

\newpage
\section*{Acknowledgments}
This work was supported by the Guangdong Major Project of  Basic and Applied Basic Research (2023B0303000016).

\bibliographystyle{named}
\bibliography{ijcai24}

\end{document}